# Spiking Sequence Machines and Transformers

Joy Bose
Independent Researcher
Bengaluru, India
*joy.bose@ieee.org*

## Abstract

Sequence learning reduces to similarity-based retrieval over a temporally indexed representation space, a constraint on any sequence model, not a property of a specific architecture. We show that a spiking Sparse Distributed Memory sequence machine (2007) and the transformer (2017) independently instantiate the same five functional operations (encoding, context maintenance, associative retrieval, storage, and decoding), with cosine similarity as the shared retrieval primitive in both. We formalise a Phase–Latency Isomorphism showing that sinusoidal positional phase and spike timing are linearly related, and prove that dot-product attention is invariant to this mapping up to a global scale factor on the positional component (Lemma 1). Empirically, frequency-compressed positional encoding fails to converge on a positionally demanding copy task, while a learned rank-based embedding matches or exceeds sinusoidal encoding, indicating that the critical property for positional representation is distance discriminability under dot-product similarity, not sinusoidal form. Time, phase, and rank are three instantiations of the same computational primitive: an ordered index whose structure survives similarity-based retrieval.

## 1. Introduction

What must any system do to learn sequences? At a minimum it must represent symbols, retain context across time, retrieve relevant associations, store learned patterns, and produce outputs. These requirements are not design choices. They follow from the structure of the prediction problem itself. A system that cannot perform any one of them cannot learn sequences in general.

If this decomposition is necessary rather than contingent, then two systems designed under radically different constraints should converge toward it independently. This paper examines two such systems.

The first is a spiking Sparse Distributed Memory sequence machine [Bose, 2007], implemented using biologically plausible leaky integrate-and-fire neurons, rank-ordered N-of-M codes, and a local Hebbian learning rule. It was motivated by computational neuroscience: how might the brain learn sequences with asynchronous, event-driven components? The second is the transformer [Vaswani et al., 2017], built on sinusoidal positional encodings, multi-head self-attention, feedforward value projection, and gradient descent at scale. It was motivated by engineering: how do we build the most capable sequence models on available hardware?

These two systems share the same five-layer functional decomposition. More specifically, the core retrieval operation, which is cosine similarity between high-dimensional vectors used for content-addressed memory lookup, is functionally identical at the level of the similarity primitive in both. Spiking SDM implementations have been empirically confirmed to achieve memory capacity comparable to non-spiking counterparts across multiple neuron models [Ajwani et al., 2021], establishing that the spiking retrieval mechanism is operationally viable.

We further show that the two systems use equivalent representations of sequential position. Sinusoidal positional phase in transformers and rank-order spike timing in spiking systems are related by a linear mapping (Proposition 1), and dot-product attention is invariant to this mapping up to a scale factor on the positional component (Lemma 1). The unifying principle is the following: any ordered index whose relative structure is preserved under inner-product similarity can serve as a positional representation. Time, phase, and rank are three instantiations of this principle.

This paper makes four contributions. First, it characterises the five functional operations that any sequence learning system must perform, and establishes that both the spiking SDM and the transformer perform all five (Section 2). Second, it formalises the Phase–Latency Isomorphism and proves attention invariance (Section 3). Third, it connects burst stability in deep spiking networks to gradient flow theory in deep learning (Section 4). Fourth, it identifies concrete implications for spiking transformer design and a falsifiable near-term experiment (Section 5).

## 2. Five Necessary Functions and Their Instantiations

We claim that sequence learning requires five functional operations, performed in order. Table 1 shows how the spiking SDM and the transformer each instantiate these operations. The following subsections examine each pair, identifying where the correspondence is exact and where it diverges. The claim is a functional necessity, since any system that generalises across arbitrary sequences must perform these operations, not that every implementation makes them architecturally explicit

| **Function** | **Spiking SDM (2007)** | **Transformer / State Space Model (SSM)** | **Match** | **Section** |
|---|---|---|---|---|
| Encoding | One-hot (1-of-A) → rank-ordered N-of-M spike burst. Geometric significance vector | Discrete token → dense d_model embedding (learned) | Functional | 2.1 |
| Context maintenance | Lambda-gated recurrent state. Scalar Λ blends shift-register and Elman recurrence | Full Key-Value (KV) cache (Transformer); input-gated state (Mamba, RWKV) | Strong / Exact for SSMs | 2.2 |
| Retrieval | SDM: cosine similarity on | Query-Key (QK) dot-product + softmax: cosine | Same similarity | 2.3 |

| Function | Spiking SDM (2007) | Transformer / State Space Model (SSM) | Match | Section |
|---|---|---|---|---|
| | significance vectors and threshold | similarity on dense embeddings | primitive. Selection mechanism differs (threshold vs. softmax) | |
| Storage / learning | Max outer-product Correlation Matrix Memory (CMM): $W_{ij} \leftarrow \max(W_{ij}, \sigma_i\sigma_j)$; local Hebbian | Feedforward Network / value projection; global stochastic gradient descent (SGD) on cross-entropy loss | Functional; learning rule diverges | 2.4 |
| Decoding | N-of-M → One-hot (1-of-A) via transposed encoder weights; Winner Take All (WTA) inhibition | Unembedding: linear projection + softmax over vocabulary | Strong | 2.5 |

*Table 1. Five necessary functions of sequence learning and their instantiation in the spiking SDM and transformer / SSM family.*

## *2.1 Encoding: Mapping Symbols into a Similarity Space*

Any sequence learning system must map discrete symbols into a space where similarity can be computed continuously. The specific mechanism matters less than the functional requirement: the resulting representation must support graded comparison between symbols. Other efficient spike encoding strategies have demonstrated this same principle in domain-specific tasks such as neuromorphic speech recognition [Yarga et al., 2023], reinforcing that multiple spiking encoding schemes can satisfy the functional requirement.

The spiking SDM represents each symbol as a rank-ordered N-of-M code: exactly N neurons fire within a burst, and their firing order is significant. The representation is a significance vector with geometrically decreasing weights:

$$s = [1, \alpha, \alpha^2, \ldots, \alpha^{(N-1)}] \qquad \alpha \text{ in } (0, 1)$$

where α is the significance ratio and components are assigned in decreasing order to the N neurons that fire. Similarity between two codewords is the normalised dot product of their significance vectors, i.e. cosine similarity. The ordered code carries substantially more information than its unordered counterpart: $\log_2(M!/(M-N)!)$ versus $\log_2(C(M,N))$ bits, a factor of 6.7 at N=255, M=256.

The transformer embedding maps each token index to a dense learned vector of dimension d_model. The two encodings differ in sparsity (N≪M active dimensions versus fully dense), learnability (fixed geometric structure versus gradient-trained), and substrate (spike timing versus floating-point arithmetic). The equivalence is functional, not parametric: both support similarity-based retrieval. A sparse structured encoding with a geometric prior and a dense learned embedding are different solutions to the same requirement.

### *2.2 Context Maintenance: Gating Past Against Present*

Sequence learning requires retention of prior inputs. Any solution must balance two competing pressures: retain enough history to capture long-range dependencies, and weight recent inputs appropriately. The design space runs from perfect unbounded recall to pure Markov processing.

The spiking SDM combines a shift register (finite-horizon, exact) and an Elman-style recurrent layer [Elman, 1990] (unbounded, approximate) using a scalar gate $\Lambda \in [0,1]$:

$$C_n = \mathrm{nofm}( \mathrm{Lambda} * \mathrm{scale}(P1 * C_{\{n-1\}}) + (1-\mathrm{Lambda}) * \mathrm{scale}(P2 * I_n))$$

where nofm selects the top m components to maintain the N-of-M code structure. At $\Lambda=0$ the context is the current input; as $\Lambda \rightarrow 1$ the context is its own history. The scalar gate implements a continuous forgetting curve.

The transformer's KV cache occupies the opposite extreme: perfect recall of all previous tokens at quadratic cost, with no decay. Selective state space models [Gu & Dao, 2024; Peng et al., 2023] occupy an intermediate position, replacing the full KV cache with input-dependent gated state updates. The spiking SDM's scalar $\Lambda$ gate is structurally consistent with this gating principle; the SSM contribution is replacing the fixed scalar with a learned, content-dependent gate. The two approaches arrived at the same design space from different directions: one from biological plausibility constraints, one from computational efficiency.

### *2.3 Retrieval: Cosine Similarity as the Shared Primitive*

The retrieval operation is where the correspondence is most precise. Both systems perform content-addressed associative retrieval: use the current state to identify which stored patterns are most relevant, and retrieve them weighted by similarity.

The spiking SDM computes the similarity between the context vector and stored memory locations using the normalised dot product of significance vectors:

$$\mathrm{sim}(X, Y) = (\mathrm{sum_i}\ X_i * Y_i) / \mathrm{sqrt}( \mathrm{sum_i}\ X_i^2 * \mathrm{sum_i}\ Y_i^2 )$$

This is cosine similarity. A threshold on this value determines which locations contribute to the retrieved output. The transformer computes the QK dot product between the query (a linear projection of the current hidden state) and all key vectors (projections of prior tokens), then applies softmax normalisation. Both use cosine similarity as the core similarity primitive for retrieval. The downstream selection mechanisms differ: hard threshold in the SDM versus softmax normalisation in the transformer. Preliminary empirical evidence suggests that spiking implementations of SDM retrieval achieve memory capacity comparable to non-spiking counterparts across LIF, Adaptive-LIF, Izhikevich [Izhikevich, 2003], and Spiking ReLU neuron models but the fundamental comparison operation is the same. This distinction matters: softmax produces a weighted combination of all values, while threshold selection produces a sparse winner-take-all output. The paper's claim of correspondence is at the level of the similarity primitive, not the full retrieval pipeline.

This equivalence is not merely structural. Preliminary empirical evidence suggests that spiking implementations of SDM retrieval achieve memory capacity comparable to non-spiking counterparts across LIF, Adaptive-LIF, Izhikevich [Izhikevich, 2003], and Spiking ReLU neuron models, across address decoder sizes from 256 to 1024 [Ajwani et al., 2021]. The spiking cosine similarity computation works in practice. Note that the Ajwani et al. implementation used unordered N-of-M codes; the capacity advantage of rank-ordered codes ($\log_2(M!/(M-N)!)$ versus $\log_2(C(M,N))$) represents an untested potential improvement.

## *2.4 Storage: Associating Context with Prediction*

Any sequence learning system must store associations between contexts and subsequent symbols. The two systems use sharply different mechanisms to do this, and this is the deepest divergence in the correspondence.

The spiking SDM uses a correlation matrix memory with a max outer-product Hebbian learning rule:

$$W_{ij} \ \texttt{<--}\ \texttt{max(}\ W_{ij},\ \ \texttt{sigma_i * sigma_j )}$$

This rule is local, online, and one-shot: no global error signal, no backward pass, no repeated presentation required. Its capacity is linear in memory size but cannot approach gradient-based learning at scale. Multiple variants are compatible with this architecture, including STDP-based [Bi & Poo, 1998] correlation memory, reservoir-based memory with Hebbian readout [Maass et al., 2002], and gated memory with neuromodulation [Bose, 2026]. Surrogate gradient methods [Neftci et al., 2019] attempt to bridge this gap by approximating the spike function as smooth during the backward pass while keeping binary spikes in the forward pass. Current spiking language models (SpikeGPT [Zhu et al., 2023], SpikeBERT [Lv et al., 2023]) operate well below transformer scale, and the learning gap remains the primary practical obstacle.

The transformer uses stochastic gradient descent with backpropagation through global cross-entropy loss. This is powerful, scalable, and biologically implausible. The functional requirement, which is to store associations between contexts and predictions, is met by both systems; the mechanism differs fundamentally.

Both systems also exhibit a form of in-context association without weight update. In the spiking SDM, eligibility traces hold significance components of a first burst temporarily in the synapses until a second burst arrives, forming a temporary association through spike-timing-dependent plasticity. In the transformer, the KV cache holds key and value vectors for all prior tokens, enabling temporary context-dependent association through attention without any parameter modification. Both are forms of fast weights.

## *2.5 Decoding: Inverting the Encoding*

Both systems invert the encoding to produce output in the original symbol space. The spiking SDM applies a transposed encoder weight matrix to the retrieved significance vector and uses winner-take-all inhibition to select the output symbol, a hard argmax. The transformer applies a learned unembedding projection followed by softmax to produce a probability distribution over the vocabulary. The functional operation is the same: map from the high-dimensional representation space back to discrete symbols. The spiking decoder is deterministic and sparse; the transformer decoder is probabilistic and dense.

## 3. The Phase–Latency Isomorphism

### *3.1 Two Representations of Sequential Position*

Beyond the five-layer correspondence, the two systems use different mechanisms to represent position within a sequence. We show these mechanisms are equivalent under a linear mapping, and that this equivalence is operationally meaningful for the attention computation.

Transformers represent position pos in a sequence using a fixed d-dimensional sinusoidal vector [Vaswani et al., 2017]:

$$PE(pos, 2i) = \sin( pos / 10000^{(2i/d)} )$$

$$PE(pos, 2i+1) = \cos( pos / 10000^{(2i/d)} )$$

where each dimension pair (2i, 2i+1) corresponds to a frequency $\omega_i = 1/10000^{(2i/d)}$. Position is encoded as phase across a bank of frequencies. The dot product PE(pos).PE(pos') depends on |pos−pos'|, making attention sensitive to positional distance.

Spiking systems encode position through spike latency within a temporal window T. The token at position pos fires at time t_spike(pos) within the window. Under a uniform latency assignment:

$$t_spike(pos) = (pos / L) * T$$

The rank of the spike within the window is identical to the positional rank. The geometric significance vector $s=[1,\alpha,...,\alpha^{(N-1)}]$ weights rank-ordered neurons in the population, so that earlier-firing (lower-position) neurons carry higher significance.

### *3.2 Proposition 1: Phase–Latency Isomorphism*

Define the frequency-scaled spike latency for frequency band i:

$$\varphi_i_spike(pos) = \omega_i * t_spike(pos) = \omega_i * (pos * T / L)$$

Substituting $\omega_i = 1/10000^{(2i/d)}$ and $\varphi_i(pos) = pos/10000^{(2i/d)}$:

$$\varphi_i_spike(pos) = (T/L) * \varphi_i(pos)$$

**Proposition 1 (Phase–Latency Isomorphism).** Under the mapping t_spike(pos) = pos*T/L, pairwise phase differences in the sinusoidal positional encoding are reproduced exactly, up to global scale factor (T/L), by pairwise differences in frequency-scaled spike latencies:

$$\varphi_i(pos) - \varphi_i(pos') = (L/T) * [ \varphi_i_spike(pos) - \varphi_i_spike(pos') ]$$

Proof: immediate from the linearity of the mapping. The scale factor (T/L) depends only on sequence length and temporal window, not on position or frequency band. The relative phase structure of the sinusoidal encoding is exactly preserved in the spike timing representation.

### *3.3 Lemma 1: Attention Invariance*

**Lemma 1 (Attention Invariance).** Let STPE(pos) = (T/L)*PE(pos) be the spike timing positional encoding derived from PE(pos) by the Phase–Latency mapping. Let e(pos) = embed(pos)+PE(pos) and e'(pos) = embed(pos)+STPE(pos) be the full token representations. For any pair (pos, pos'), the positional component of the attention logit under STPE differs from that under PE by a global scale factor $(T/L)^2$. After softmax normalisation, the relative ordering of attention weights across positions is preserved, provided that the positional and content components of the logit are comparably scaled.

The reasoning is as follows. Under the Phase–Latency mapping, STPE = (T/L) · PE, so the positional component of the attention logit scales by $(T/L)^2$. After softmax, attention weights depend only on the ranking of logits across positions, not their absolute values, and ranking is preserved under any positive monotone scaling. The scale factor therefore acts as an inverse softmax temperature on the positional component, sharpening positional attention at longer sequences, without changing which positions receive the highest weight. The invariance holds in regimes where positional information contributes meaningfully to the attention logit; if content contributions dominate, the mapping has negligible practical effect in either direction. The linearity of the mapping makes the isomorphism operational: it is precisely this property that yields attention invariance. Lemma 1 is operationally significant, not merely formal. It states that the computational operation which consumes positional encodings, dot-product attention, produces the same relative outputs under spike timing encoding as under sinusoidal encoding. The Phase–Latency Isomorphism is therefore not a curiosity about proportional numbers; it is a statement about what the attention mechanism can and cannot distinguish.

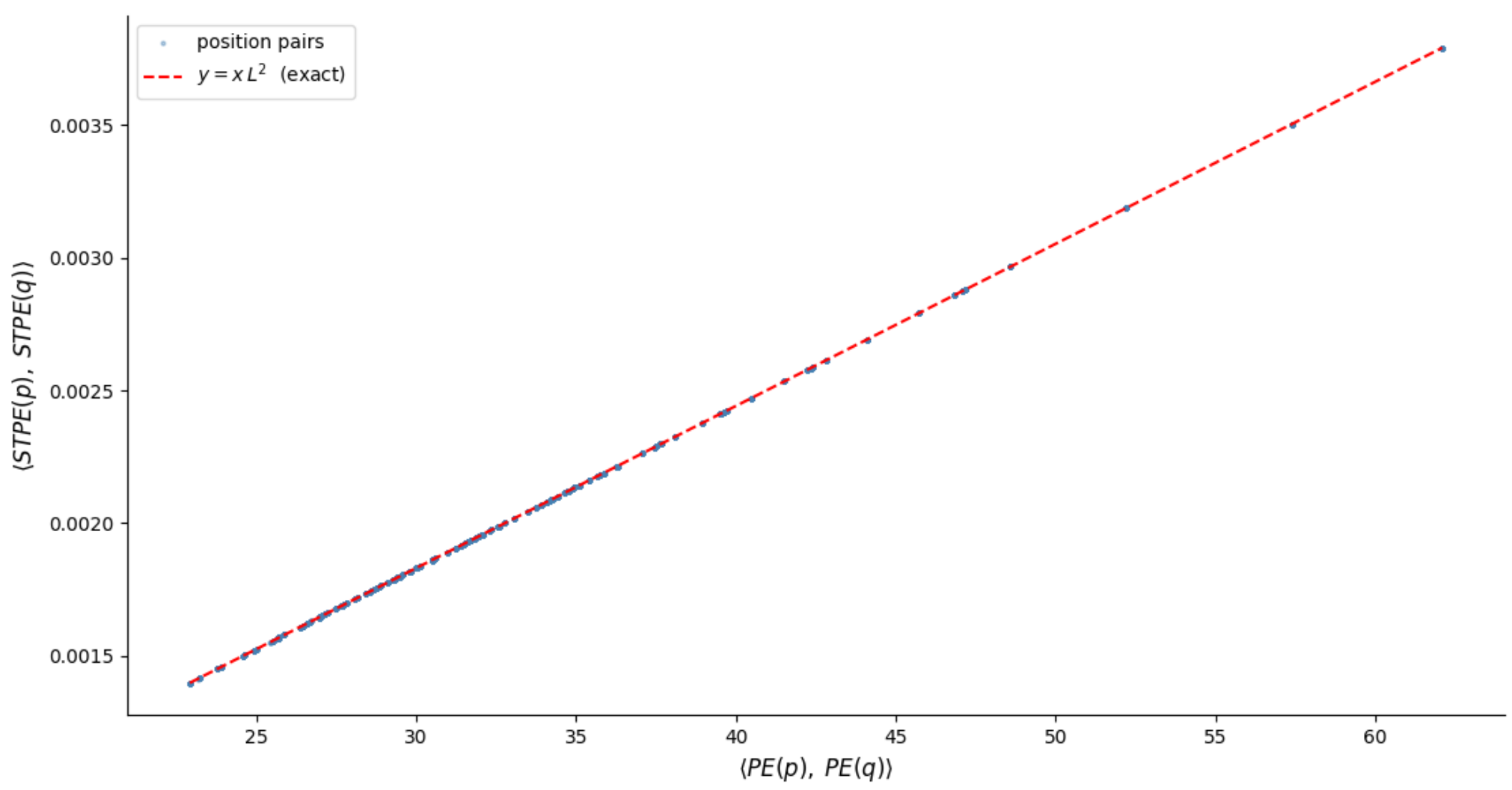

(a)

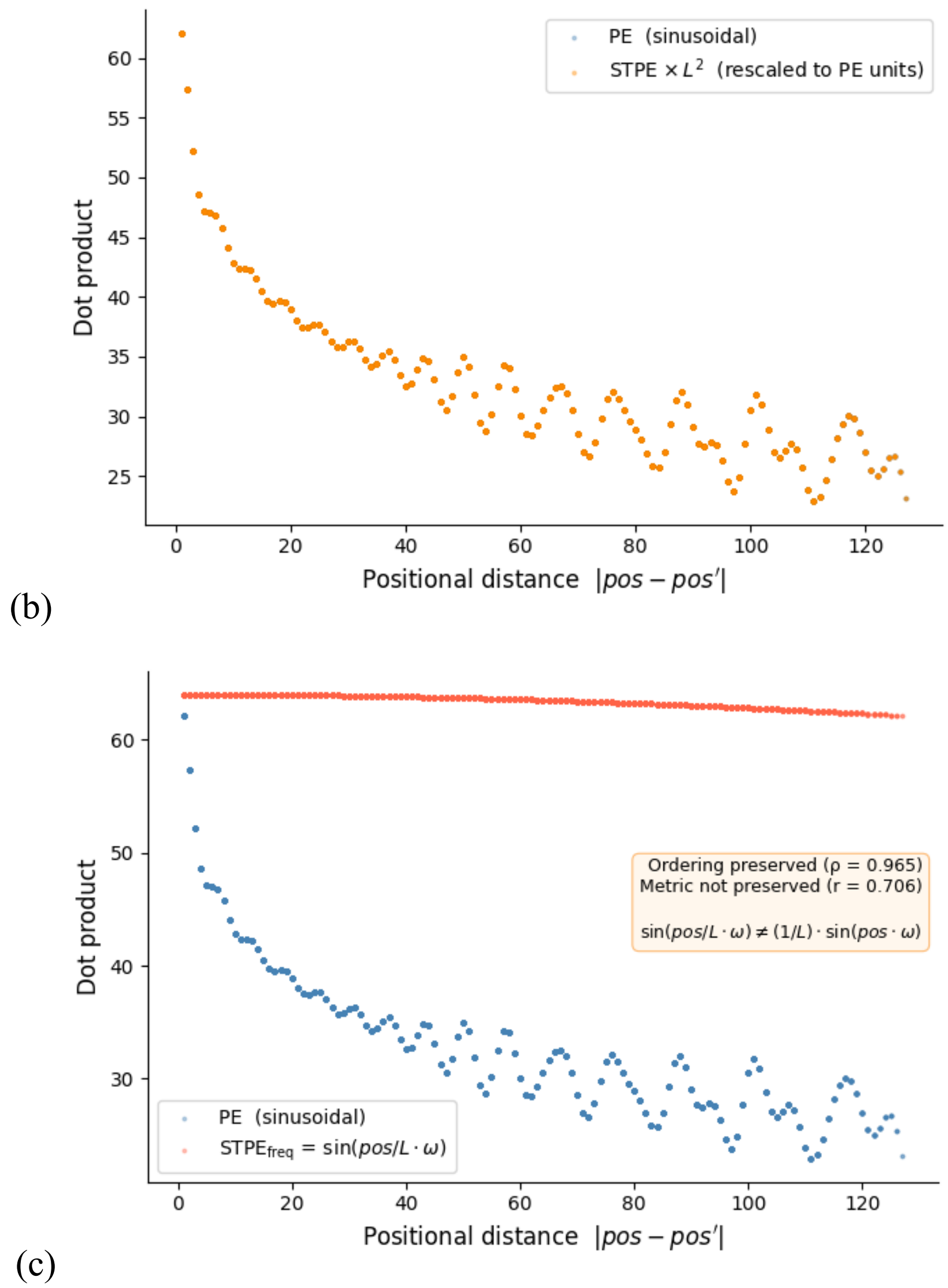

*Figure 1. Empirical validation of the Phase–Latency Isomorphism (Proposition 1). L=128, d=128. Left: scatter of ⟨PE(p), PE(q)⟩ versus ⟨STPE(p), STPE(q)⟩ across all position pairs — points lie exactly on $y = x/L^2$, confirming ⟨STPE, STPE⟩ = $(T/L)^2$⟨PE, PE⟩. Right: pairwise dot products as a function of positional distance |pos − pos′|, comparing PE (blue) and amplitude-scaled STPE rescaled to PE units (orange). Pearson r = 1.000, Spearman ρ = 1.000.*

## *3.4 The Unifying Principle*

Proposition 1 and Lemma 1 together establish a triangle of equivalent representations:

```
position in sequence <—> sinusoidal phase <—> spike latency <—> ordinal rank
```

These are not three different solutions to the problem of positional representation. They are three instantiations of a single computational primitive: an ordered index whose relative structure survives inner-product similarity computation. The claim is not that any ordered index works: it must be one whose pairwise differences are preserved under the dot product used for retrieval. Proposition 1 shows that spike latency satisfies this requirement. The rank-ordered significance vector with geometric weighting satisfies it by construction, since the cosine similarity of significance vectors privileges agreement in early (high-rank) positions.

## 4. Burst Stability and Gradient Flow: A Structural Parallel

A persistent problem in deep network theory is signal propagation: activations either collapse to zero or diverge to infinity as they pass through many layers, making learning impossible. Modern solutions include residual connections, layer normalisation, and initialisation schemes calibrated to maintain signal variance across layers [Schoenholz et al., 2017; Yang & Schoenholz, 2017]. The same structural problem appears in deep spiking networks, and an empirical solution was identified in the 2007 sequence machine literature, independently and earlier.

### *4.1 The Phase Transition in Spiking Networks*

In networks of rank-ordered N-of-M spiking neurons, spike bursts must propagate from layer to layer without fading or exploding, and without accumulating temporal dispersion that destroys rank-order information. Bose (2007) studied this empirically in feedforward networks of up to 200 RDLIF layers with 256 neurons per layer and 10% connectivity, following an 11-of-256 code.

The central finding was a sharp phase transition in the neural threshold. Below a critical value, spike activity grows from layer to layer and destroys the rank-order code. Above it, activity dies within a few layers. At the critical threshold (between 87.693 and 87.694 in the experimental configuration), burst temporal dispersion stabilises to a bounded range across 100 layers, regardless of initial dispersion. This was verified across 48 independent runs. Dispersion decreases monotonically with connectivity, approaching zero at full connectivity. The system operates at the edge of chaos: the boundary between order and disorder that the biological neural network literature had long associated with computational competence.

### *4.2 The Structural Parallel*

The connection to modern gradient flow theory is direct. The vanishing/exploding gradient problem has the same root cause: signal either collapses or diverges as it passes through many layers. Schoenholz et al. (2017) and Yang & Schoenholz (2017) characterise the ordered-chaotic phase boundary in deep random networks analytically, showing that trainability requires operating near this boundary. The 2007 empirical result identifies the same boundary in a spiking network, via a different quantity (burst dispersion rather than gradient variance) but the same structural phenomenon.

The solutions are also structurally parallel. The 2007 solution was feedback reset inhibition: an inhibitory neuron fires once N output spikes have occurred in a layer, resetting all neurons and bounding the burst. This is a per-layer activation constraint, structurally analogous to layer normalisation in transformers, which bounds activation magnitude at each layer. In both cases the constraint is local (per-layer rather than global) and must bound activation from above and below.

This parallel extends the correspondence of Section 2 to the training dynamics of deep networks. It is, to our knowledge, not previously noted in the spiking network literature.

# 5. Implications

## *5.1 On the Necessity of the Decomposition*

The five-layer decomposition identified in Section 2 is not arbitrary. If the argument of Section 1 is correct, that these five operations are necessary for any sequence learning system, then their convergent appearance in two systems designed under radically different constraints is expected, not coincidental. The differences between the systems (sparse versus dense encoding, decaying versus perfect recall, local versus global learning) are differences in how each operation is implemented, not in whether it is performed.

The interesting question is not whether these systems are similar but why the same operation, which is cosine similarity for retrieval, emerges as the solution in both cases. One answer is that cosine similarity is the natural inner product on the unit sphere, and that representing symbols as normalised vectors (whether sparse significance vectors or dense embeddings) makes cosine similarity the canonical comparison operation. The functional constraint (support graded retrieval) combined with the representational choice (high-dimensional normalised vectors) uniquely determines the operation.

## *5.2 Spiking Attention: A Design Proposal*

The structural equivalence of Section 2.3 suggests a concrete design for spiking attention. The QK dot product is replaced by an SDM address decoder computing cosine similarity on rank-ordered N-of-M codes. The softmax over all key vectors is replaced by winner-take-all inhibition: the most-similar address decoder neurons fire first and suppress the rest through feedback reset inhibition, the same mechanism identified in Section 4.1 as the spiking analogue of layer normalisation.

Spikformer [Zhou et al., 2022] implements spiking attention using binary spike codes. Rank-ordered N-of-M codes carry higher information content ($\log_2(M!/(M-N)!)$ versus $\log_2(C(M,N))$ for binary unordered codes) and offer the significance ratio $\alpha$ as a tunable parameter controlling information density versus noise robustness. Extending the Ajwani et al. (2021) Nengo implementation to rank-ordered codes and measuring the capacity improvement is a well-defined next experiment.

## *5.3 A Falsifiable Near-Term Experiment*

Lemma 1 makes a falsifiable prediction. If spike timing encoding and sinusoidal encoding are equivalent under attention, then a model trained with rank-normalised positional encoding PE’(pos,i) = sin((pos/L)*ω_i) should achieve comparable sequence learning performance to one trained with standard sinusoidal PE. The rank-only variant, a learnable embedding indexed purely by position rank, without sinusoidal structure, tests the stronger claim that ordinal information alone is sufficient. The experiment reveals a more nuanced result than simple confirmation: frequency-compressed encoding fails to converge on a positionally demanding task, while the pure rank embedding outperforms sinusoidal PE, sharpening the theoretical claim in an unexpected direction.

| **Encoding** | **Final BPC** | **Does it Converge** |
|---|---|---|
| Sinusoidal PE (baseline) | 0.000308 | Yes (~1000 steps) |
| Freq-compressed PE | 0.090129 | No (plateau) |
| Pure rank embedding (learned) | 0.000003 | Yes (~500 steps) |

*Table 2. Copy task BPC at 10,000 steps (L=64, d=64, V=50, seed=42)*

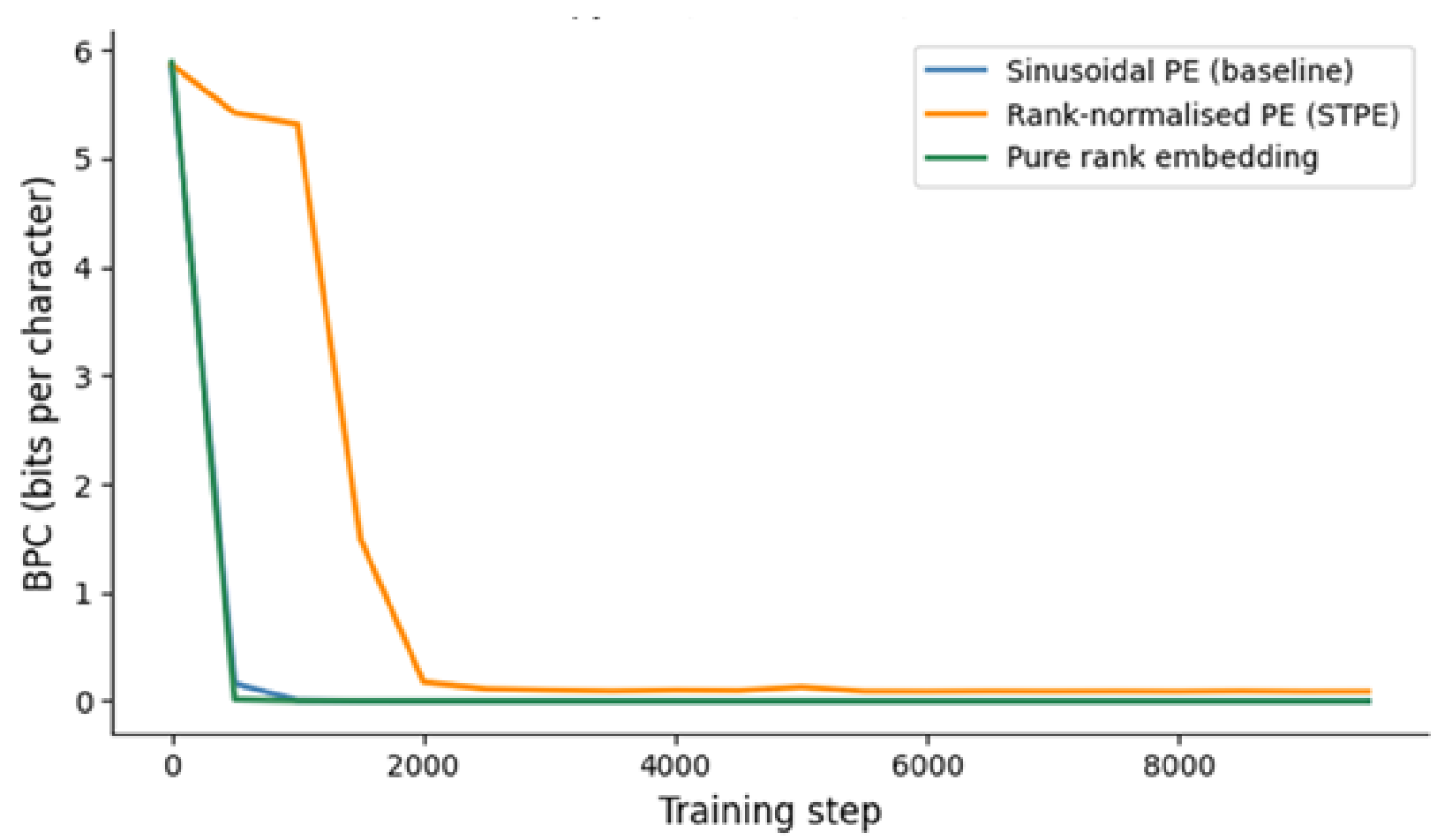

*Figure 2. Learning curves (BPC vs training step) for three positional encodings on the copy task (L=64, d=64, V=50). Sinusoidal PE (blue) converges by step ~1000. Pure rank embedding (green) converges faster and reaches lower final BPC. Frequency-compressed PE (orange) fails to converge, consistent with the loss of positional distance information shown in Figure 1(c). The result confirms that ordinal structure is sufficient for positional learning, and that the critical property is distance discriminability under dot-product similarity, not sinusoidal form.*

Table 1 reports final BPC after 10,000 steps on the copy task (L=64, d=64, V=50). Sinusoidal PE converges to near-zero BPC by step 1000. The frequency-compressed encoding ($\sin((pos/L)\cdot\omega_i)$) fails to converge, reaching a plateau of BPC ≈ 0.090, a qualitative failure consistent with the loss of positional distance information demonstrated in Figure 1(b,c). This confirms that frequency compression is a distinct encoding that does not satisfy Lemma 1, and that the Phase–Latency Isomorphism requires amplitude scaling, not frequency compression. Frequency compression collapses the phase range such that dot-product similarity becomes weakly dependent on positional distance (Figure 1c), preventing attention from distinguishing near from distant positions, which is a structural failure, not a convergence speed issue. This does not contradict Lemma 1: the invariance applies to amplitude scaling (STPE = $(T/L)\cdot$PE), not to frequency transformations such as $\sin((pos/L)\cdot\omega_i)$, which alter the similarity structure as shown in Figure 1. Most significantly, the pure rank embedding (a learned nn.Embedding indexed by position) converges faster than sinusoidal PE and reaches lower final BPC (0.000003 vs 0.000308), suggesting that learned ordinal representations can be competitive with, or in some cases more efficient than, fixed sinusoidal encodings, though this finding is on a single synthetic task and warrants replication at larger scale and across multiple seeds. Results are from a single random seed (42); variability across seeds has not yet been characterised, and these findings should be treated as preliminary.

### *5.4 Limitations*

- The five-function decomposition of Section 2 is argued by necessity, not proved formally. A tighter treatment would characterise the minimal set of operations required by sequence prediction and prove completeness.
- Lemma 1 holds for the positional component of the attention logit and requires that positional and content contributions are comparably scaled. In models with very high-dimensional content embeddings or learned PE, the positional component may be negligible, limiting the scope of the invariance.
- The Phase–Latency Isomorphism assumes a linear, uniform latency mapping. Non-linear or content-dependent mappings (RoPE [Su et al., 2021], ALiBi [Press et al., 2022], learned PE) may carry more information and are not covered.
- The Ajwani et al. (2021) empirical confirmation used unordered N-of-M codes. The capacity advantage of rank-ordered codes claimed in Section 2.3 has not been directly tested in a spiking implementation.
- The burst stability result was obtained on networks of up to 200 layers with 256 neurons per layer. Generalisation to transformer-scale spiking networks is an open question.

## 6. Related Work

Spiking transformers have been approached from the hardware efficiency direction. Spikformer [Zhou et al., 2022] replaces dense attention with binary spike-based attention and achieves competitive image classification results while reducing energy cost on neuromorphic hardware. SpikeGPT [Zhu et al., 2023] and SpikeBERT extend this to language modelling. These works carry over sinusoidal positional encodings unchanged. The present paper addresses the complementary question: what is the natural positional representation for a spiking system, and how does it relate to sinusoidal encoding?

Ellwood (2024) demonstrates that short-term Hebbian learning can implement transformer-like attention through a match-and-control principle, providing complementary mechanistic evidence for the functional convergence identified here. Recent work on relative positional encoding in spiking transformers has shown that spike timing differences can encode relative positions with practical energy efficiency gains on neuromorphic hardware. The present paper's contribution relative to these works is the formal Phase–Latency Isomorphism (Proposition 1 and Lemma 1), the systematic five-layer correspondence as a necessary decomposition argument, and the empirical finding that frequency-compressed encoding destroys distance discriminability while learned rank embeddings remain effective.

Selective state space models [Gu & Dao, 2024; Peng et al., 2023] arrived at gated recurrent context from the deep learning direction. The gating principle they instantiate is structurally consistent with scalar-gated context models in the spiking literature; the contribution of SSMs is replacing the fixed scalar with an input-dependent learned gate.

Rank-order coding [Thorpe & Gautrais, 1998; VanRullen & Thorpe, 2002] provides the biological and computational motivation for spike latency as an information carrier. The connection between rank-order coding and positional encoding in transformers, formalised here as the Phase–Latency Isomorphism, is to our knowledge novel to this paper.

Sparse Distributed Memory [Kanerva, 1988] and its N-of-M implementations [Furber et al., 2007] form the memory substrate of the spiking system. The functional connection between SDM retrieval (cosine similarity with threshold) and transformer attention (cosine similarity with softmax) has been noted informally; the present paper formalises it and situates it within the broader five-layer correspondence.

Surrogate gradient methods [Neftci et al., 2019; Bellec et al., 2020] attempt to train spiking networks with backpropagation using smooth spike function approximations in the backward pass. These methods are compatible with the architecture described here as an alternative to the Hebbian learning rule, and represent the most promising route to closing the learning gap identified in Section 2.4.

## 7. Conclusion

Sequence learning requires similarity-based retrieval over a temporally indexed representation space. This is not a description of one architecture but a constraint on any architecture that learns sequences. Two systems developed independently under radically different constraints, a spiking SDM sequence machine and a transformer, both satisfy this constraint and both implement the same five-layer functional decomposition as a consequence.

The retrieval operation is identical in both systems: cosine similarity between high-dimensional vectors for content-addressed memory lookup. The positional representations are equivalent under a linear mapping (Proposition 1), and dot-product attention is invariant to this mapping in regimes where positional information contributes meaningfully to the logit (Lemma 1). The burst stability phase transition in deep spiking networks is structurally parallel to the vanishing/exploding gradient problem in deep learning, and the per-layer feedback inhibition solution is structurally parallel to layer normalisation.

Time, phase, and rank are three instantiations of the same computational primitive: an ordered index whose relative structure survives inner-product similarity computation. The rank-only ablation described in Section 5.3 is a direct falsifiable test of this claim, runnable in PyTorch in a day. If ordinal structure alone is sufficient for positional encoding, the unifying principle proposed here has practical implications for both spiking and non-spiking sequence models.

The failure of frequency-compressed encoding and the success of pure rank embedding together sharpen the paper's central claim: what survives similarity-based retrieval is not sinusoidal structure specifically, but ordinal structure, and when that structure is learned rather than fixed, it is not merely sufficient but can be more efficient than the hand-crafted sinusoidal prior. These results suggest that positional encodings must preserve not only ordering but sufficient dispersion in similarity space to remain discriminable under dot-product attention, a principle that holds across both spiking and transformer architectures.

## References

Ajwani, R. D., Lalan, A., Bhattacharya, B. S., & Bose, J. (2021). Sparse Distributed Memory using Spiking Neural Networks on Nengo. Bernstein Conference 2021. arXiv:2109.03111.

Bellec, G., Scherr, F., Subramoney, A., Hajek, E., Salaj, D., Legenstein, R., & Maass, W. (2020). A solution to the learning dilemma for recurrent networks of spiking neurons. Nature Communications, 11(1), 3625.

Bi, G., & Poo, M. (1998). Synaptic modifications in cultured hippocampal neurons: Dependence on spike timing, synaptic strength, and postsynaptic cell type. Journal of Neuroscience, 18(24), 10464–10472.

Bose, J. (2007). Engineering a Sequence Machine Through Spiking Neurons Employing Rank-Order Codes. PhD thesis, University of Manchester.

Bose, J. (2026). What I Built in 2007, and Why It Looks a Bit Like a Transformer. Medium, March 17, 2026. https://joyboseroy.medium.com/what-i-built-in-2007-and-why-it-looks-a-bit-like-a-transformer-dbf3683a0ebe

Ellwood, I. (2024). Short-term Hebbian learning can implement transformer-like attention. PLOS Computational Biology, 20(1), e1011843. arXiv:2310.19812.

Elman, J. L. (1990). Finding structure in time. Cognitive Science, 14(2), 179–211.

Furber, S. B., Brown, G., Bose, J., Cumpstey, J. M., Marshall, P., & Shapiro, J. L. (2007). Sparse distributed memory using rank-order neural codes. IEEE Transactions on Neural Networks, 18(3), 648–659.

Gu, A., & Dao, T. (2024). Mamba: Linear-time sequence modeling with selective state spaces. First Conference on Language Modeling.

Izhikevich, E. M. (2003). Simple model of spiking neurons. IEEE Transactions on Neural Networks, 14(6), 1569–1572.

Kanerva, P. (1988). Sparse Distributed Memory. MIT Press.

Lv, C., Xu, W., Zheng, Q., & others. (2023). SpikeBERT: A language spikformer learned from BERT with knowledge distillation. arXiv:2308.06259.

Maass, W., Natschläger, T., & Markram, H. (2002). Real-time computing without stable states: A new framework for neural computation based on perturbations. Neural Computation, 14(11), 2531–2560.

Neftci, E. O., Mostafa, H., & Zenke, F. (2019). Surrogate gradient learning in spiking neural networks. IEEE Signal Processing Magazine, 36(6), 51–63.

Peng, B., et al. (2023). RWKV: Reinventing RNNs for the Transformer Era. arXiv:2305.13048.

Press, O., Smith, N. A., & Lewis, M. (2022). Train short, test long: Attention with linear biases enables input length extrapolation. ICLR 2022.

Schoenholz, S. S., Gilmer, J., Ganguli, S., & Sohl-Dickstein, J. (2017). Deep information propagation. ICLR 2017.

Su, J., Lu, Y., Pan, S., Murtadha, A., Wen, B., & Liu, Y. (2021). RoFormer: Enhanced transformer with rotary position embedding. arXiv:2104.09864.

Thorpe, S., & Gautrais, J. (1998). Rank order coding. Computational Neuroscience: Trends in Research, 113–118.

VanRullen, R., & Thorpe, S. J. (2002). Surfing a spike wave down the ventral stream. Vision Research, 42(23), 2593–2615.

Vaswani, A., Shazeer, N., Parmar, N., Uszkoreit, J., Jones, L., Gomez, A. N., Kaiser, L., & Polosukhin, I. (2017). Attention is all you need. NeurIPS, 30.

Yang, G., & Schoenholz, S. S. (2017). Mean field residual networks: On the edge of chaos. NeurIPS 2017.

Yarga, S. F., Rouat, J., & Wood, S. U. N. (2023). Efficient spike encoding algorithms for neuromorphic speech recognition. Proceedings of the International Conference on Neuromorphic Systems (ICONS).

Zhou, Z., Zhu, Y., He, C., Wang, Y., Yan, S., Tian, Y., & Yuan, L. (2022). Spikformer: When spiking neural network meets transformer. arXiv:2209.15425.

Zhu, R. J., Zhao, Q., Li, G., & Eshraghian, J. K. (2023). SpikeGPT: Generative pre-trained language model with spiking neural networks. arXiv:2302.13939.